\pdfoutput=1

\documentclass[11pt]{article}

\usepackage[review]{acl}

\usepackage{times}
\usepackage{latexsym}
\usepackage{amsmath}
\usepackage{amsfonts}
\usepackage{amssymb}
\usepackage{graphicx}
\usepackage{epstopdf}
\usepackage{tikz}
\usepackage[utf8]{inputenc}
\usepackage{CJKutf8}
\usepackage{multirow}
\usepackage[normalem]{ulem}
\useunder{\uline}{\ul}{}
\usepackage{tikz}
\usetikzlibrary{positioning} 
\usepackage{xcolor}
\usepackage{pgfplots}

\usepackage[T1]{fontenc}

\usepackage[utf8]{inputenc}

\usepackage{microtype}

\title{Learning Disentangled Semantic Representations for Zero-Shot Cross-Lingual Transfer in Multilingual Machine Reading Comprehension}

\author{Linjuan Wu$^{1}$\footnotemark[1] , {\bf Shaojuan Wu$^{1}$\footnotemark[1]} , {\bf Xiaowang Zhang$^{1,2}$}, {\bf Deyi Xiong$^{1}$\footnotemark[2]} , {\bf Shizhan Chen$^{1}$\footnotemark[2]} , \\ {\bf  Zhiqiang Zhuang$^1$}, {\bf Zhiyong Feng$^1$}\\
        $^1$College of Intelligence and Computing, Tianjin University, Tianjin, China, 300350 \\ $^2$Tianjin University-Aishu Data Intelligence Joint Laboratory, Tianjin, China \\ \texttt{Tianjin\_Research@aishu.cn}\\
        \texttt{\{wulinjuan1997,shaojuanwu,dyxiong,shizhan\}@tju.edu.cn}}

\begin{document}
\maketitle

\renewcommand{\thefootnote}{\fnsymbol{footnote}}
\footnotetext[1]{These authors contributed equally to this work and should be considered co-first authors.}
\footnotetext[2]{Corresponding authors.}

\begin{abstract}
Multilingual pre-trained models are able to zero-shot transfer knowledge from rich-resource to low-resource languages in machine reading comprehension (MRC). However, inherent linguistic discrepancies in different languages could make answer spans predicted by zero-shot transfer violate syntactic constraints of the target language. In this paper, we propose a novel multilingual MRC framework equipped with a Siamese Semantic Disentanglement Model (S$^2$DM) to disassociate semantics from syntax in representations learned by multilingual pre-trained models. To explicitly transfer only semantic knowledge to the target language, we propose two groups of losses tailored for semantic and syntactic encoding and disentanglement. Experimental results on three multilingual MRC datasets (i.e., XQuAD, MLQA, and TyDi QA) demonstrate the effectiveness of our proposed approach over models based on mBERT and XLM-100.

\end{abstract}

\section{Introduction}\label{introduction}
Multilingual pre-trained language models (PLMs) \cite{BERT, XLM, XLM-R} have been widely explored in cross-lingual understanding tasks. However, zero-shot transfer method based on multilingual PLMs does not work well for low-resource language MRC. Such multilingual MRC models could roughly detect answer spans but may fail to predict the precise boundaries of answers \citep{YuanSBGLDFJ20}. 

\begin{figure}
\centering
\includegraphics[width=7.5cm]{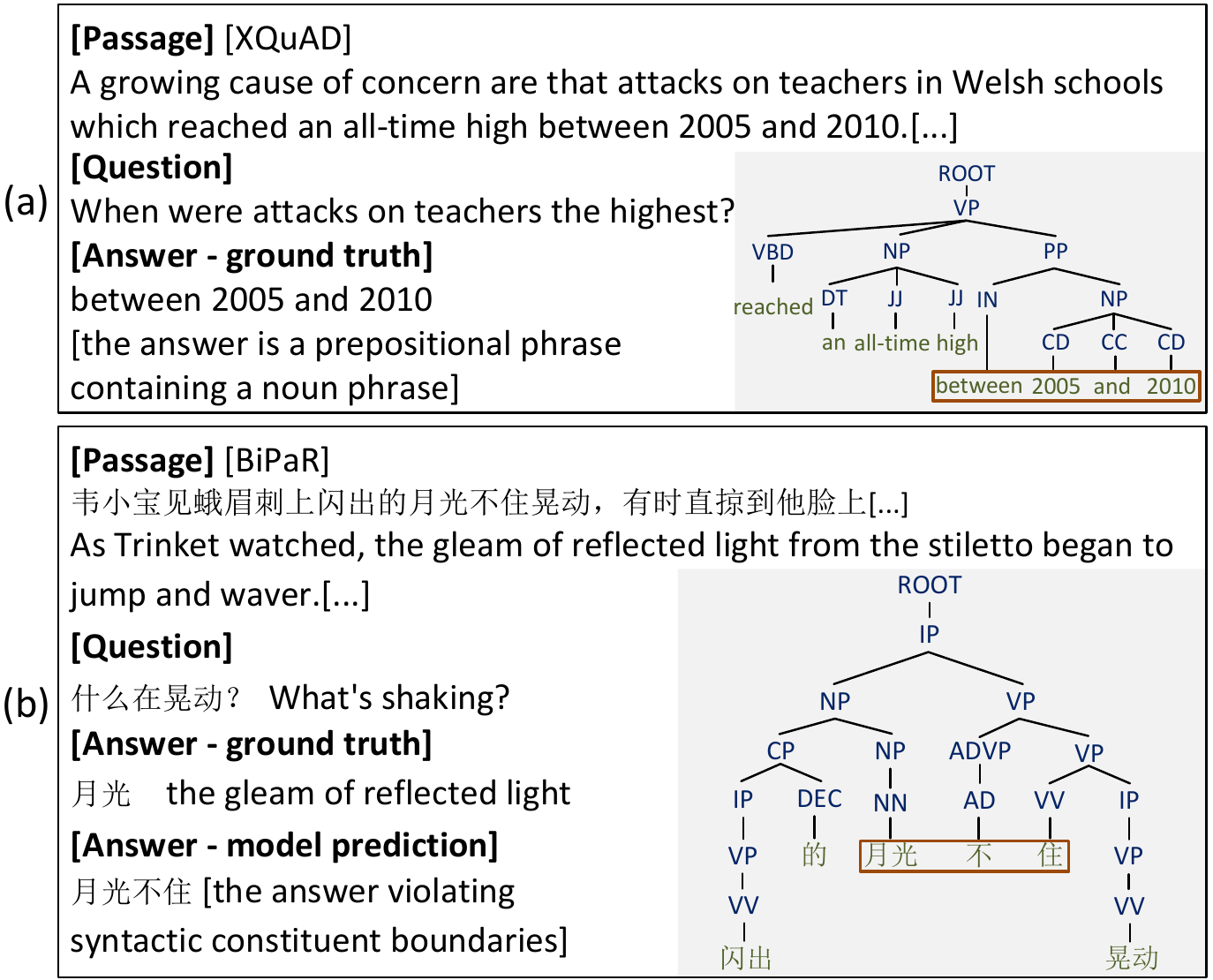}
\caption{Relations between answer spans and syntactic constituents. (a) An example from XQuAD~\citep{XQuAD} where the ground-truth answer is a syntactic constituent. (b) A case from BiPaR~\citep{BiPaR} where the answer predicted by a model transferred from English to Chinese violates syntactic constituent boundaries in the target language. }
\label{example}
\end{figure}


In order to address this issue, existing methods mainly resort to external resources. Based on the finding that ~70\% of answer spans are language-specific phrases (e.g., named entities, noun phrases) in MLQA \citep{MLQA}, \citet{YuanSBGLDFJ20} propose an additional language-specific knowledge phrase masking (LAKM) task to enhance boundary detection for low-resource languages. \citet{CalibreNet} present a boundary calibration model stacked over a base sequence labeling module, introducing a phrase boundary recovery task to pre-train the calibration module on large-scale multilingual datasets synthesized from Wikipedia documents. These two methods rely on external resources, which are not always easily available.

As illustrated in Figure~\ref{example}(b), the transfer model may violate syntactic constraints for answer spans in the target language (e.g., the predicted answer \begin{CJK}{UTF8}{gbsn}"月光不住"\end{CJK} crossing the boundaries of two subtrees). An intuitive assumption is that the majority of answer spans respect syntactic constituency boundaries (i.e., syntactic constraint, illustrated by the case in Figure~\ref{example}(a)). On four multilingual MRC datasets, we use Stanford CoreNLP\renewcommand{\thefootnote}{\arabic{footnote}}\footnote{\url{https://stanfordnlp.github.io/CoreNLP/}} to collect syntax parse trees and calculate the percentages of ground-truth answers that respect syntactic constituent boundaries. As shown in Table~\ref{statistic}, over 87\% of answer spans respect the syntactic constraint.

On the bilingual parallel MRC corpus BiPaR \citep{BiPaR}, we have compared two MRC models: a monolingual MRC model trained on the Chinese data of BiPaR vs. an mBERT-based MRC model trained on the English data of BiPaR and adapted to Chinese via zero-shot transfer. For questions where the monolingual model correctly predicts the answer and respect syntactic constraint, 23.15\% of them are incorrectly predicted by the transfer model, and the predicted answers violate the syntactic constraint, illustrated by the case in Figure~\ref{example}(b). This suggests that the source language syntax may have a negative impact on the answer boundary detection in the target language during zero-shot transfer, due to the linguistic discrepancies between the two languages. 

\begin{table}
\centering
\resizebox{7.6cm}{!}{
\begin{tabular}{l|c|c|c|c}
\hline
                  		  & XQuAD                 & MLQA                 & TyDi QA-GoldP        & BiPaR              \\ \hline
English               & 89.08\%                 & 90.11\%              & 89.12\%                     & 90.99\%                   \\ 
Chinese             & 88.05\%                 & 87.57\%              & -                                 & 95.73\%         \\ \hline
\end{tabular}}
\caption{ The percentages of answer spans that respect syntactic constituent boundaries in four multilingual MRC datasets in both English and Chinese. }
\label{statistic}
\end{table}

\begin{figure*}
\centering
\includegraphics[width=15cm]{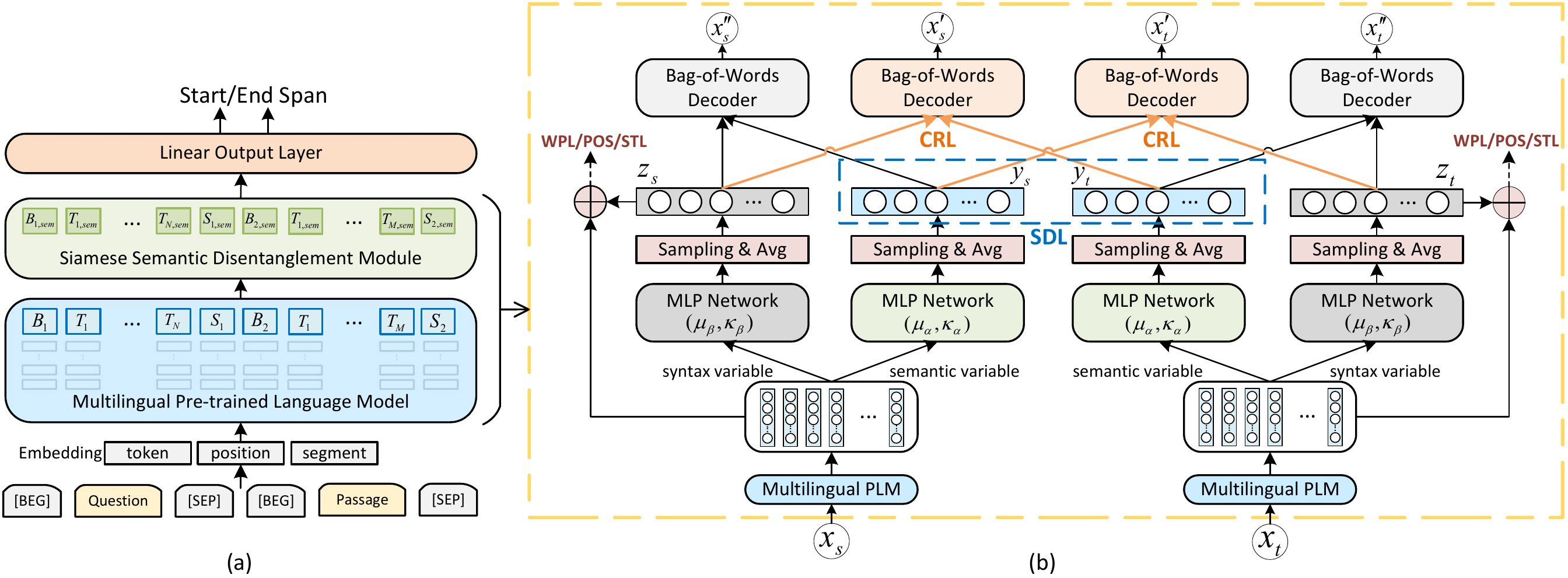}
\caption{Diagram of the proposed zero-shot cross-lingual transfer framework for multilingual MRC. (a) The overview of our multilingual MRC framework. (b) The architecture of S$^2$DM that is composed of two siamese networks with shared parameters for the source and target language. Once trained, only the output of source language MLP network is fed into the linear output layer. The subscripts $s$ and $t$ in (b) represent the source and target language respectively. CRL: cross-lingual reconstruction loss. SDL: semantic discrimination loss. WPL: word position loss. POS: Part-of-Speech loss. STL: syntax tree loss. }
\label{whole_model}
\end{figure*}

However, linguistic discrepancies are diverse and it is difficult to learn them. We hence propose to decouple semantics from syntax in pre-trained models for multilingual MRC, transforming the learning of linguistic discrepancies into universal semantic information. Specifically, we propose a Siamese Semantic Disentanglement Model (S$^2$DM) that utilises two latent variables to learn semantic and syntactic vectors in multilingual pre-trained representations. As shown in Figure~\ref{whole_model}(a), stacking a linear output layer for MRC over the disentangled semantic representation layer, we can fine-tune the multilingual PLMs on the rich-resource source language and transfer only disentangled semantic knowledge into the target language MRC. Our model aims to reduce the negative impact of the source language syntax on answer boundary detection in the target language.

To disassociate semantic and syntactic information in PLMs well, we introduce objective functions of learning cross-lingual reconstruction and semantic discrimination together with losses of incorporating word order information and syntax structure information (Part-of-Speech tags and syntax parse trees). We use a publicly available multilingual sentence-level parallel corpus with syntactic labels to train S$^2$DM.


To summarize, our main contributions are as follows. 
\begin{itemize}
  \item We propose a multilingual MRC framework that explicitly transfers semantic knowledge of the source language to the target language to reduce the negative impact of source syntax on answer span detection in the target language MRC.  
  \item We propose a siamese semantic disentanglement model that can effectively separate semantic from syntactic information of multilingual PLMs with semantics/syntax-oriented losses. 
  \item Experimental results on three multilingual MRC datasets ( XQuAD, MLQA, and TyDi QA) demonstrate that our model can achieve significant improvements of 3.13 and 2.53 EM points over two strong baselines, respectively.
\end{itemize}

\section{Related Work}

\paragraph{Cross-lingual/Multilingual Machine Reading Comprehension} Recent advances in multilingual MRC evaluation datasets \cite{XQuAD, MLQA, TyDiQA} trigger research interests in multilingual and cross-lingual MRC \cite{HsuLL19, Dual-BERT, YuanSBGLDFJ20, LiuSPGYJ20, self-training, 210705002}. \citet{HsuLL19} investigate cross-lingual transfer capability of multilingual BERT (mBERT) on MRC tasks and find that zero-shot learning based on PLM is feasible, even between distant languages, such as English and Chinese. Various approaches have been proposed on top of multilingual MRC based on PLMs. \citet{Dual-BERT} propose a method that combines multilingual BERT and back-translation for cross-lingual MRC. In order to effectively leverage translation data and reduce the impact of noise in translations, \citet{LiuSPGYJ20} propose a cross-lingual training approach based on knowledge distillation for multilingual MRC. \citet{YuanSBGLDFJ20} present two auxiliary tasks: mixMRC and LAKM to introduce additional phrase boundary supervision into the fine-tuning stage. \citet{CalibreNet} propose a pre-trained boundary calibration module based on the output of a base zero-shot transfer model, refining the boundaries of initial answers.  

Different from the above studies, we mainly consider the impact of syntactic divergences between the source and target language in zero-shot cross-lingual transfer based on multilingual PLMs, and attempt to disassociate semantics from syntax and only transfer semantics to the target language. 

\paragraph{Disentangled Representation Learning}
Recently, there has been a growing amount of work on learning disentangled latent representations in NLP tasks \cite{ZhangYYSC19,HuYLSX17,NeubigZYH18}. In this aspect, the most related work to our syntax-semantics decoupling method is the vMF-Gaussian Variational Autoencoder (VGVAE) model proposed by \citet{ChenTWG19}. It is a generative model using two latent variables to represent semantics and syntax of the sentence, developed for monolingual setting and trained with paraphrases. It uses paraphrase reconstruction loss and a discriminative paraphrase loss to learn semantic representations and word order information for syntactic representations. We adapt this model to multilingual syntax-semantics disentanglement. We use bilingual sentence pairs to train our model with a cross-lingual reconstruction loss and semantic discrimination loss. To better disentangle semantics from complex and diverse syntax in multilingual PLMs, we introduce two additional syntax-related losses for incorporating POS tags and syntax trees.

\section{Approach}

Figure~\ref{whole_model} shows the architecture of our multilingual MRC framework with the proposed siamese semantic disentanglement model.

\subsection{Multilingual MRC Framework}
Our multilingual MRC framework consists of three essential components: the multilingual PLM layer, the siamese semantic disentanglement module, and the linear output layer. The output representations from the multilingual PLM are fed into S$^2$DM to disassociate semantic and syntactic information. Only the disentangled semantic representations are input to the linear output layer for predicting answer spans in passages.

In order to facilitate the zero-shot cross-lingual transfer of only semantic knowledge from the rich-resource source language to the low-resource target language, we take a two-stage training strategy. First, we pre-train S$^2$DM with parallel data (see Section~\ref{ssdm}) while the parameters of the multilingual PLM are frozen. Once S$^2$DM is trained, only the output of source language MLP network is fed into the linear output layer for MRC. In the second step, we freeze the parameters of the S$^2$DM and fine-tune the entire multilingual MRC framework on MRC data of the source language. 

\subsection{Siamese Semantic Disentanglement Model} \label{ssdm}
In S$^2$DM, we assume that a sentence $x$ is generated by a semantic and syntactic variable, i.e., $y$ and $z$, independently. We follow VGVAE~\citet{ChenTWG19} to use the von Mises-Fisher (vMF) distribution for the semantic variable and the Gaussian distribution for the syntactic variable. Formally, the joint probability of the sentence and its two latent variables can be factorized as:
\begin{equation}
\setlength{\abovedisplayskip}{1pt}
\setlength{\belowdisplayskip}{1pt}
\begin{aligned}
p_{\theta}(x, y, z)&=p_{\theta}(y)p_{\theta}(z)p_{\theta}(x|y,z)
\end{aligned}
\end{equation}
where $p_{\theta}(x|y,z)$ is a generative model consisting of bag-of-words decoder.

The variational inference process of VGVAE uses a factorized approximated posterior $q_{\phi}(y|x)q_{\phi}(z|x)=q_{\phi}(y,z|x)$ with the objective function that maximizes a lower bound of the marginal log-likelihood:
\begin{equation}
\setlength{\abovedisplayskip}{1pt}
\setlength{\belowdisplayskip}{1pt}
\begin{aligned}
L_{VGVAE}&=L_{RL} + {\rm KL}(q_{\phi}(z|x)||p_{\theta}(z))  \\ & + {\rm KL}(q_{\phi}(y|x)||p_{\theta}(y)),
\end{aligned}
\label{vgvae}
\end{equation}
\setlength{\abovedisplayskip}{1pt}
\setlength{\belowdisplayskip}{1pt}
\begin{equation}
L_{RL}  = \mathbb{E}_{y \sim q_{\phi}(y|x) \atop z \sim q_{\phi}(z|x)} \big{[}-{\rm log}\ p_{\theta}(x|y,z)\big{]}
\label{rl}
\end{equation}
where $q_{\phi}(y|x)$ is subject to vMF$(\mu_{\alpha }(x), \kappa_{\alpha }(x))$ while $q_{\phi}(z|x)$ follows $N(\mu_{\beta}(x), diag(\kappa_{\beta}(x)))$. The prior $p_{\theta}(y)$ and $p_{\theta}(z)$ follows the uniform distribution vMF$(\cdot, 0)$ and a standard Gaussian distribution respectively. Eq.(\ref{rl}) is the \textbf{reconstruction loss (RL)} of the generator. In our model, we adopt a multilayer perceptron (MLP) network to learn the mean ($\mu$) and variance ($\kappa$) of two distributions. As pre-trained representations are contextually-encoded token vectors, latent variable vectors obtained by sampling from the distributions need to be averaged so as to output sentence-level semantic and syntactic vector. 

Since S$^2$DM uses a Siamese network for both the source and target language, the disentanglement between semantics and syntax is conducted for the two languages simultaneously with two parameter-shared subnetworks, as shown in Figure~\ref{whole_model}(b). 

We attempt to extract rich semantic information from multilingual representations which is universal for multiple languages and contains less syntactic information. Except for the conventional reconstruction loss, we propose two additional losses on parallel data to encourage the latent variable $y$ to capture semantic information: a \textbf{Cross-lingual Reconstruction Loss (CRL)} and \textbf{Semantic Discrimination Loss (SDL)}. The former estimates the cross-entropy loss when we use the semantic representation $y_t$ of the target language to reconstruct the source input and use the source semantic representation $y_s$ for target reconstruction. The latter is used to force the learned source semantic representation $y_s$ to be as close as possible to the target semantic representation $y_t$ since the semantic meanings of the parallel source and target sentence is equivalent to each other. The two losses are estimated as follows:
\begin{equation}
\setlength{\abovedisplayskip}{1pt}
\setlength{\belowdisplayskip}{1pt}
\begin{aligned}
L_{CRL} &= \mathbb{E}_{y_{t} \sim q_{\phi}(y|x_{t}) \atop z_{s} \sim q_{\phi}(z|x_{s})} \big{[}-{\rm log}\ p_{\theta}(x_{s}|y_{t}, z_{s})\big{]}\\
&+\mathbb{E}_{y_{s} \sim q_{\phi}(y|x_{s}) \atop z_{t} \sim q_{\phi}(z|x_{t})} \big{[}-{\rm log}\ p_{\theta}(x_{t}|y_{s}, z_{t})\big{]},
\end{aligned}
\label{crl}
\end{equation}
\setlength{\abovedisplayskip}{1pt}
\setlength{\belowdisplayskip}{1pt}
\begin{equation}
\begin{aligned}
L_{SDL} &= {\rm max}\big{\{}0, \delta - {\rm sim}(y_{s}, y_{t}) + {\rm sim}(y_{s}, n_{t})\big{\}}\\
&+{\rm max}\big{\{}0, \delta - {\rm sim}(y_{s}, y_{t}) + {\rm sim}(n_{s}, y_{t})\big{\}}
\end{aligned}
\label{sdl}
\end{equation}
where ${\rm sim}(\cdot, \cdot)$ is a cosine similarity score function. The margin $\delta$ is a hyperparameter to control the gap between parallel sentence pair $(y_{s}, y_{t})$ and two non-parallel sentence pairs $(y_{s}, n_{t})$ and $(n_{s}, y_{t})$. $n_{s}$ is the semantic vector of a negative sample, which has the highest cosine similarity to $y_{s}$. Specially, as partial sentences in our corpus are parallel in more than two languages, we limit the data range of negative sampling to only 2-way parallel pairs. $n_{t}$ are obtained in the similar way to $n_{s}$. 

In order to guide S$^2$DM to disassociate syntactic information into the syntactic latent variable $z$, we also define three losses tailored for capturing different types of syntactic information. First, we employ \textbf{Word Position Loss (WPL)} , defined as follows:
\begin{equation}
\setlength{\abovedisplayskip}{1pt}
\setlength{\belowdisplayskip}{1pt}
L_{WPL}=\mathbb{E}_{z \sim q_{\phi}(z|x)} \big{[}-\sum_{i} {\rm log}\ {\rm softmax}(f(h_i))_i\big{]},
\end{equation}
where ${\rm softmax}(\cdot)_i$ indicates the probability of the $i$th word at position $i$, and $f(\cdot)$ is a three-layer feedforward neural network with input $h_i = [e_{i};z]$ that is the concatenation of the syntactic variable $z$ and the embedding vector $e_i$ of the multilingual PLM for the $i$th token in the input sentence.

In addition, we define a Part-of-Speech and syntax tree loss to encourage S$^2$DM to isolate deeper syntactic information from pre-trained representations. POS tagging is a sequence labeling task,  which can be regarded as a multi-class classification problem for each token in a sentence. Hence, we define \textbf{Part-of-Speech (POS) Loss} as a cross-entropy style loss as follows:
\begin{equation}
\setlength{\abovedisplayskip}{1pt}
\setlength{\belowdisplayskip}{1pt}
L_{POS} = \sum_{i}\big{[}\sum_{j=1}^{m}-{\rm log}\ {\rm softmax}(g(h_i))_{j=class}\big{]}
\end{equation}
where $g(\cdot)$ is a linear layer, softmax$(\cdot)_{j=class}$ estimates the probability of gold POS tag $class$, $m$ is the number of different POS tags.

For learning structural information, we design \textbf{Syntax Tree Loss (STL)}. Many studies have found that PLMs can encode syntactic structures of sentences (\citet{structural_probe, multiprobe}). Inspired by \citet{structural_probe}, we formulate syntactic parsing from pre-trained word representations as two independent tasks: depth prediction of a word and distance prediction of two words in the parse tree. Given a matrix $B \in \mathbb{R}^{k \times m}$ as a linear transformation, the losses of these two subtasks are defined as:
\begin{equation}
\setlength{\abovedisplayskip}{3pt}
\setlength{\belowdisplayskip}{3pt}
L_{depth} = \sum_{i}(\left \| w_i \right \| - \left \| Bh_i \right \|_2^2),
\end{equation}
\begin{equation}
\setlength{\abovedisplayskip}{3pt}
\setlength{\belowdisplayskip}{3pt}
L_{distance} = \sum_{i,j}\big{|}{\rm d}_{T}(w_{i}, w_{j})-{\rm d}_{B}(h_{i}, h_{j})\big{|}   
\end{equation}
where $\left \| w_i \right \|$ is the parse depth of a word defined as the number of edges from the root of the parse tree to $w_i$, and  $\left \| Bh_i \right \|_2$ is the tree depth L2 norm of the vector space under the linear transformation. ${\rm d}_{T}(w_{i}, w_{j})$ is the number of edges in the path between the $i$th and $j$th word in the parse tree $T$. As for ${\rm d}_{B}(h_{i}, h_{j})$, it can be defined as the squared $L_2$ distance after transformation by B:
\begin{equation}
{\rm d}_{B}(h_{i}, h_{j}) = (B(h_{i}-h_{j}))^T(B(h_{i}-h_{j})) 
\end{equation}

To induce parse trees, we minimize the summation of the above two losses $L_{depth}$ and $L_{distance}$, and $L_{STL}$ is defined as:
\begin{equation}
\setlength{\abovedisplayskip}{3pt}
\setlength{\belowdisplayskip}{3pt}
L_{STL} = L_{depth} + L_{distance} 
\end{equation}

According to the different syntactic tasks, we train two S$^2$DM variants: S$^2$DM\_POS and S$^2$DM\_SP (SP for syntactic parsing), where their training objectives are defined as follows:
\begin{equation}
\setlength{\abovedisplayskip}{3pt}
\setlength{\belowdisplayskip}{3pt}
L_{1} = L_{VGVAE} + L_{CRL} + L_{SDL} + L_{WPL} + L_{POS},\nonumber 
\end{equation}
\begin{equation}
L_{2} = L_{VGVAE} + L_{CRL} + L_{SDL} + L_{WPL} + L_{STL}\nonumber 
\end{equation}

\subsection{Generalization Analysis} \label{ta}
In this section, we analyze the generalization of our decoupling-based multilingual MRC model. 

By two reconstruction losses Eq.(\ref{rl}) and Eq.(\ref{crl}), we will prove that the syntactic and semantic vectors obtained by S$^2$DM are language-agnostic. Since the mathematic structures of Eq.(\ref{rl}) and Eq.(\ref{crl}) are the same, we take one part of Eq.(\ref{crl}) for analysis. Due to $z_s$ and $y_t$ are independent of each other, $p_{\theta}(x_{s},z_{s}|y_{t})=p_{\theta}(x_{s},z_{s})$. We obtain:
\begin{equation}
\begin{aligned}
&\ \mathbb{E}_{y_{t} \sim q_{\phi}(y|x_{t}) \atop z_{s} \sim q_{\phi}(z|x_{s})} \big{[}-{\rm log}\ p_{\theta}(x_{s}|y_{t}, z_{s})\big{]}\\
&=\mathbb{E}_{y_{t} \sim q_{\phi}(y|x_{t})} \big{(}\sum_{ z_{s} \sim q_{\phi}(z|x_{s})} p_{\theta}(z_s){\rm log}\frac{p_{\theta}(z_s)}{p_{\theta}(x_{s},z_{s}|y_{t})}\big{)}\\
&={\rm KL}(p_{\theta}(z_s)||p_{\theta}(x_{s},z_{s})) \nonumber 
\end{aligned}
\end{equation}
 Similarly,
 
\begin{small}
\begin{equation}
\begin{aligned}
\mathbb{E}_{y_{s} \sim q_{\phi}(y|x_{s}) \atop z_{t} \sim q_{\phi}(z|x_{t})} [-{\rm log}\ p_{\theta}(x_{t}|y_{s}, z_{t})]= {\rm KL}(p_{\theta}(z_t)||p_{\theta}(x_{t},z_{t})) \nonumber 
\end{aligned}
\end{equation}
\begin{equation}
L_{RL}  = {\rm KL}(p_{\theta}(y_s)||p_{\theta}(x_{s},y_{s}))+{\rm KL}(p_{\theta}(y_t)||p_{\theta}(x_{t},y_{t})) \nonumber 
\end{equation}
\end{small}

Minimizing ${\rm KL}(q_{\phi}(z|x)||p_{\theta}(z))$ and ${\rm KL}(q_{\phi}(y|x)||p_{\theta}(y))$ will eventually fit both $p_{\theta}(x_{s},z_{s})$ and $p_{\theta}(x_{t},z_{t})$ into the same distribution. In the same way, both $p_{\theta}(x_{s},y_{s})$ and $p_{\theta}(x_{t},y_{t})$ also fit to the same distribution, no matter what the target language is. This is consistent with our motivation to use the siamese network.

Furthermore, the semantic discrimination loss in Eq.(\ref{sdl}) guarantees that the semantic vectors of the source language and the target language are similar to each other. Minimizing Eq.(\ref{sdl}) can be equivalent to:
\begin{equation}
\setlength{\abovedisplayskip}{3pt}
\setlength{\belowdisplayskip}{3pt}
\left\{
\begin{array}{rl}
{\rm sim}(y_{s}, y_{t}) > {\rm sim}(y_{s}, n_{t}) + \delta \\
{\rm sim}(y_{s}, y_{t}) >{\rm sim}(n_{s}, y_{t}) + \delta \nonumber 
\end{array}
\right.
\end{equation}
which is to maximize ${\rm sim}(y_{s}, y_{t})$ to encourages the target semantic vector to approach parallel source semantic vector. 

In summary, S$^2$DM can obtain language-agnostic semantic and syntactic vectors. Therefore, our multilingual MRC model is suitable even for low-resource languages without training data for the decoupling model.

\section{Experiment}

\subsection{Datasets}
To verify the effectiveness of our multilingual MRC model, we conducted experiments on three multilingual question answering benchmarks: 

\textbf{XQuAD} \citep{XQuAD} consists of 11 datasets of different languages translated from the SQuAD v1.1 ~\citep{SQuAD1.0} development set, including Spanish (es), German (de), Greek (el), Russian (ru), Turkish (tr), Arabic (ar), Vietnamese (vi), Thai (th), Chinese (zh), Hindi (hi), and Romanian (ro). 

\textbf{MLQA} \citep{MLQA} consists of over 5K extractive MRC instances in 7 languages: English (en), Arabic (ar), German (de), Spanish (es), Hindi (hi), Vietnamese (vi) and Chinese (zh). MLQA is also highly parallel, with MRC instances parallel across 4 different languages on average.

\textbf{TyDi QA-GoldP} is the gold passage task in TyDi QA \citep{TyDiQA} covering 9 typologically diverse languages: Arabic (ar), Bengali (bg), English (en), Finnish (fi), Indonesian (id), Korean (ko), Russian (ru), Swahili (sw), Telugu (te). It is a more challenging MRC benchmark as questions have been written without seeing the answers, leading to 3 and 2 times less lexical overlap than XQuAD and MLQA, respectively \citep{XTREME}.

\subsection{Baseline Models}
We used the following two multilingual PLMs to build our MRC model to conduct experiments:

\textbf{mBERT} is the multilingual version of BERT \citet{BERT}, with 177M parameters, is pre-trained on the Wikipedia of 104 languages to optimize the masked language modeling objective.

\textbf{XLM-100} uses a pre-training objective similar to that of mBERT but with a larger number of parameters (578M) and a larger shared vocabulary than mBERT, and is trained on the same Wikipedia data covering 100 languages as mBERT. 

Furthermore, we compared with a strong baseline that uses external knowledge to enhance cross-lingual MRC:

\textbf{LAKM} is a pre-trained task proposed in \cite{YuanSBGLDFJ20} by introducing external sources for phrase-level masked language modeling task. The external corpus contain 363.5k passages and 534k knowledge phrases in four languages: English (en), French (fr), German (de), and Spanish (es).

\subsection{Setup}
For S$^2$DM, we collected approximately 26k labelled parallel sentence pairs from the Universal Dependencies (UD 2.7) Corpus \citep{UD2.7} as the training set. The training set covers 20 languages and overlap with 13 languages of three MRC datasets. We used Universal POS tags and HEAD tags in UD 2.7 for the POS tagging and syntactic parsing task. We chose data from the Chinese semantic textual similarity (STS) task \citep{STS} as the development set. For hyper-parameters in S$^2$DM, the learning rate was set to 5e-5, the margin $\delta$ was 0.4, and the latent variable dimensions was 200.

For our multilingual MRC models and two baseline models, we fine-tuned them on the SQuAD v1.1 \citep{SQuAD1.0} and evaluated them on the test data of the three multilingual MRC datasets. For models based on mBERT, we fine-tuned them for 3 epochs with a training batch size of 32 and a learning rate of 2e-5. We fine-tuned models based on XLM-100 for 2 epochs with a training batch size of 16 and a learning rate of 3e-5.

\subsection{Experiment Results}
The overall experimental results are shown in Table~\ref{all}. All our tests were conducted under the conditions of zero-shot transfer. Our models (S$^2$DM\_POS, S$^2$DM\_SP combined with XLM-100 or mBERT) significantly outperform both XLM-100 and mBERT baselines on three datasets. S$^2$DM\_SP achieves the best performance, indicating that the learning of deeper syntax information is compelling. Especially, compared with baselines on the TyDi QA-Gold dataset, S$^2$DM\_SP based on XLM-100 and mBERT gains 4.1\%, 4.2\% EM improvements on average across 9 languages, respectively.

\begin{table}
\centering
\resizebox{7.5cm}{!}{
\begin{tabular}{ll|c|c|c|c|c|c}
\hline
\multirow{2}{*}{} & & \multicolumn{2}{c|}{XQuAD}      & \multicolumn{2}{c|}{MLQA}       & \multicolumn{2}{c}{TyDi QA}    \\ \cline{3-8} 
                  &                                             	& EM             	  	& F1                 		& EM             		& F1             		& EM             		& F1              \\ \hline
                  &XLM-100                             	& 45.3          	  	& 70.9         		& 38.5          		& 66.4          		& 33.4         		& 61.7          \\ 
XLM-100  &XLM+S$^2$DM\_POS     	&46.6           	  	&72.7           		&40.1        		&67.3           		& 35.7          		& 63.6\\
                  &XLM+S$^2$DM\_SP         	& \textbf{47.7}  	& \textbf{73.5 }  	& \textbf{ 41.4 } 	& \textbf{68.9} 	& \textbf{37.5} 	& \textbf{65.5}      \\  \hline
              &mBERT                                		& 48.5       		& 63.3          		& 41.2          		& 58.5          		& 43.6           		& 57.6          \\ 
mBERT&mBERT+S$^2$DM\_POS    	& 49.4          		& 63.7          		& 42.8          		& 59.9          		& 46.3          		& 58.7          \\
              &mBERT+S$^2$DM\_SP      	& \textbf{49.8} 	& \textbf{64.1} 	& \textbf{43.3} 	& \textbf{60.3} 	&\textbf{47.8}   &\textbf{60.1}      \\  \hline

\end{tabular}}
\caption{The average experimental results on XQuAD, MLQA and TyDi QA dataset. }
\label{all}
\end{table}

\begin{table*}
\resizebox{\textwidth}{!}{
\begin{tabular}{ll|cccccccccccc|c}
\hline
\multicolumn{2}{c|}{}                   & \multicolumn{13}{c}{XQuAD (EM/F1)}                                                                                                                                                                                          \\ \cline{3-15} 
\multicolumn{1}{c}{\multirow{-2}{*}{}} & 	& en                                 		& ar                                 	& de                                 & el                                 & es                                 & hi                                 & ro                                 & ru                                 & th                                 & tr                                 & vi                                 & zh                &avg                 \\ \hline
          		&XLM-100                       	&{66.5}/{86.5}  &{35.6}/{72.4}  &{53.8}/{80.9}  &{37.9}/{66.3}  &{54.6}/{81.0}  &{39.9}/{64.9}  &{56.6}/{79.6}  &{54.0}/\textbf{79.5}  &{10.3}/{27.0}  &{42.0}/{72.4}  &{49.5}/{75.4}  &{42.7}/{65.4}  &{45.3}/{70.9} \\

XLM-100	&XLM+S$^2$DM\_POS	&67.5/87.4					&\textbf{40.2}/\textbf{74.9} 		&54.2/80.8 &41.9/71.3 &55.4/82.1 &40.0/66.2 &56.4/79.6 &54.0/79.3 &\textbf{13.8}/\textbf{38.9} &41.9/70.8 &50.6/75.8 &42.9/65.1 &46.6/72.7 \\

			&XLM+S$^2$DM\_SP	&\textbf{68.3}/\textbf{88.0} 	&39.8/\textbf{74.9}    			&\textbf{55.8}/\textbf{81.7} &\textbf{44.1}/\textbf{72.4} &\textbf{56.8}/\textbf{82.5} &\textbf{40.5}/\textbf{66.5} &\textbf{59.0}/\textbf{81.7} &\textbf{54.2}/\textbf{79.5} &13.3/38.3 &\textbf{44.5}/\textbf{72.9}   &\textbf{51.3}/\textbf{76.1 } &\textbf{44.5}/\textbf{67.6} &\textbf{47.7}/\textbf{73.5} \\

\hline
                 &mBERT                                 	&{72.6}/{83.6}  &{44.3}/\textbf{60.6}  &{54.0}/{69.6}  &{46.0}/{61.1}  &{57.3}/{74.9}  &{38.3}/{53.3}  &{58.3}/{72.5}  &{54.0}/{69.6}  &{30.9}/{39.9}  &{33.8}/{50.9}  &{46.1}/{65.9}  &{46.3}/{57.4}  &{48.5}/{63.3}\\ 
mBERT   &mBERT +S$^2$DM\_POS  	&\textbf{73.4}/{83.2}  &\textbf{44.9}/{59.9}  &\textbf{55.6}/\textbf{71.9}  &{44.8}/{59.7}  &\textbf{57.4}/\textbf{75.0}  &{41.3}/{55.7}  &{58.1}/{72.4}  &\textbf{55.3}/\textbf{71.2}  &\textbf{32.7}/\textbf{40.7}  &{34.0}/{50.8}  &{48.2}/{67.4}  &{47.1}/{56.9}  &{49.4}/{63.7}\\
                &mBERT +S$^2$DM\_SP      	&{73.2}/\textbf{84.0}  &{43.3}/{60.0}  &{55.2}/{70.7}  &\textbf{46.6}/\textbf{61.8}  &{57.1}/{74.1}  &\textbf{42.7}/\textbf{56.5}  &\textbf{59.5}/\textbf{73.4}  &{54.6}/{70.3}  &{30.4}/{38.9}  &\textbf{36.3}/\textbf{51.4}  &\textbf{49.8}/\textbf{69.7}  &\textbf{48.9}/\textbf{58.5}  &\textbf{49.8}/\textbf{64.1} \\  \hline

 \multicolumn{2}{c|}{}              & \multicolumn{13}{c}{MLQA (EM/F1)}      \\ \hline                                                                                                                                                                                   
			&XLM-100       			&{59.1}/{81.8}  &{27.0}/{62.8}  &{43.5}/{71.3} &-    &{42.7}/{73.8}  &{29.3}/{56.4} & -& -& -& - &{37.4}/{65.0}  &{30.1}/{53.7}  &{38.5}/{66.4}\\
XLM-100   &XLM+S$^2$DM\_POS 	&\textbf{61.1}/82.8  &30.5/65.7 &43.9/71.2  &-   &43.1/73.5 &31.5/58.0 & -& -& -& - &39.7/66.7 &30.5/53.1 &40.1/67.3 \\
               	&XLM+S$^2$DM\_SP 	&\textbf{61.1}/\textbf{83.0} &\textbf{31.2}/\textbf{67.1} &\textbf{45.9}/\textbf{72.9} &-   &\textbf{43.6}/\textbf{74.1} &\textbf{34.1}/\textbf{61.2} & -& -& -& - &\textbf{41.4}/\textbf{68.5} &\textbf{32.3}/\textbf{55.6} &\textbf{41.4}/\textbf{68.9} \\ 
\hline
               &mBERT                                  	&{67.0}/{79.3}  &{31.5}/{49.5}  &{43.8}/{58.3} &-    &{45.8}/{64.1}  &{29.4}/{45.2} & -& -& -& - &{37.5}/{57.3}  &{34.5}/{56.1}  &{41.2}/{58.5}\\ 
               &LAKM                                  	&{66.8}/\textbf{80.0}  &-  &\textbf{45.5}/\textbf{60.5} &-    &\textbf{48.0}/\textbf{65.9}  &- & -& -& -& - &-  &-  &-\\
mBERT  &mBERT+S$^2$DM\_POS  	&{66.3}/{79.5}  &\textbf{32.4}/{50.2}  &{45.1}/{59.7} &-    &{46.8}/{65.1}  &{30.8}/{46.0} & -& -& -& - &{39.5}/{59.4}  &\textbf{38.4}/{59.1}  &{42.8}/{59.9}\\
               &mBERT+S$^2$DM\_SP   		&\textbf{67.5}/{79.8}  &{32.1}/\textbf{50.5}  &{45.3}/{59.9} &-    &{47.2}/{65.0}  &\textbf{32.0}/\textbf{46.9} & -& -& -& - &\textbf{41.1}/\textbf{60.6}  &{38.0}/\textbf{59.3}  &\textbf{43.3}/\textbf{60.3}\\  
\hline
\end{tabular}}
\caption{EM and F1 score of 12 languages on the XQuAD and MLQA dataset.}
\label{xquad}
\end{table*}

\begin{table*}
\centering
\resizebox{\textwidth}{!}{
\begin{tabular}{ll|ccccccccc|c}
\hline
\multicolumn{2}{c|}{\multirow{2}{*}{}} & \multicolumn{10}{c}{TyDi QA-GoldP (EM/F1)}                              \\ \cline{3-12} 
\multicolumn{1}{c}{}      & & en        & ar          &bg          & fi         & id         & ko         & ru         & sw         & te     &avg   \\ \hline
			&XLM-100                        		&{52.9}/{78.1}  &{31.1}/{69.8}  &{29.2}/{57.7}  &{39.3}/{65.3}  &{42.8}/{69.0}  &{1.4}/{24.9}  &{36.8}/{70.2}  &{32.9}/{59.2}  &{34.4}/{61.1}  &{33.4}/{61.7}\\
XLM-100   &XLM\_S$^2$DM\_POS  	&{52.3}/{76.1} &{30.4}/{69.5} &\textbf{37.2}/{66.1} &{37.5}/{64.6} &{44.1}/{68.4} &\textbf{1.8}/{25.3} &{39.4}/{72.4} &\textbf{41.9}/{62.7} &{37.1}/\textbf{67.4} &{35.7}/{63.6}  \\
          		&XLM\_S$^2$DM\_SP  		&\textbf{53.6}/\textbf{78.5} &\textbf{34.4}/\textbf{72.3} &33.6/\textbf{66.8} & \textbf{47.7}/\textbf{72.7} &\textbf{45.5}/\textbf{69.4} &1.5/\textbf{28.8} &\textbf{46.3}/\textbf{75.6} &37.5/\textbf{63.1} &\textbf{37.2}/62.6 &\textbf{37.5}/\textbf{65.5}\\ 
\hline
              	&mBERT                          		&{65.5}/{75.3}  &{43.8}/{59.5}  &{39.8}/{54.9}  &{44.0}/\textbf{56.9}  &{45.3}/{59.8}  &{41.7}/{49.8}  &{41.4}/{64.4}  &{32.3}/{50.0}  &{39.0}/{48.2}  &{43.6}/{57.6}\\ 
mBERT	&mBERT+S$^2$DM\_POS	&\textbf{66.1}/{74.8}  &{44.2}/\textbf{60.9}  &{41.6}/{53.3}  &{41.9}/{55.6}  &{46.5}/{60.2}  &{45.3}/{51.7}  &{42.9}/{63.6}  &{43.3}/{55.8}  &\textbf{44.8}/\textbf{52.7}  &{46.3}/{58.7} \\
             	&mBERT+S$^2$DM\_SP 	&{65.9}/\textbf{76.6}  &\textbf{44.7}/{60.7}  &\textbf{44.2}/\textbf{55.2}  &\textbf{45.1}/{56.5}  &\textbf{47.3}/\textbf{60.9}  &\textbf{48.2}/\textbf{55.0}  &\textbf{44.3}/\textbf{65.5}  &\textbf{45.9}/\textbf{58.1}  &{44.4}/{52.0}  &\textbf{47.8}/\textbf{60.1}\\ 

\hline

\end{tabular}}
\caption{EM and F1 score of 9 languages on the TyDi QA-GoldP dataset.}
\label{tydiqa}
\end{table*}

The results of 12 languages in XQuAD and MLQA are shown in Table~\ref{xquad}. For cross-lingual transfer performance, our models are better than the two baselines in terms of either EM or F1 on all 11 low-resource target languages. On the MLQA dataset, LAKM uses a larger extra corpus to train a better backbone language model, while our method with less external data can still achieve similar performance in German (de) and Spanish (es). 

The TyDi QA-GoldP dataset is more challenging than XQuAD and MLQA. The results of TyDi QA-GoldP are shown in Table~\ref{tydiqa}, and our models are superior to the baselines in terms of either EM or F1 for all 8 low-resource target languages. Significantly, XLM+S$^2$DM\_SP outperforms the XLM-100 baselines by 8.4\%, 9.5\% in EM for Finnish (fi), Russian(ru), respectively. The language families of these two languages are different from that of English. The evaluation results on these three datasets verify the effectiveness of our proposed method.


In Section~\ref{ta}, we theoretically analyze the generalization of our model. The results on the three datasets show the effectiveness on five languages not included in the training target languages for S$^2$DM. The five languages are Romanian (ro), Vietnamese (vi) in XQuAD and Bengali (bg), Swahili (sw), Telugu (te) in TyDi QA-GoldP, which are resource-scarce and have different language families from English. Significantly, mBERT+S$^2$DM\_SP outperforms the mBERT baseline by 13.6\% in EM for Swahili (sw).

\section{Analysis}
\subsection{Ablation Study}

\begin{figure}
\centering
\includegraphics[width=7.5cm]{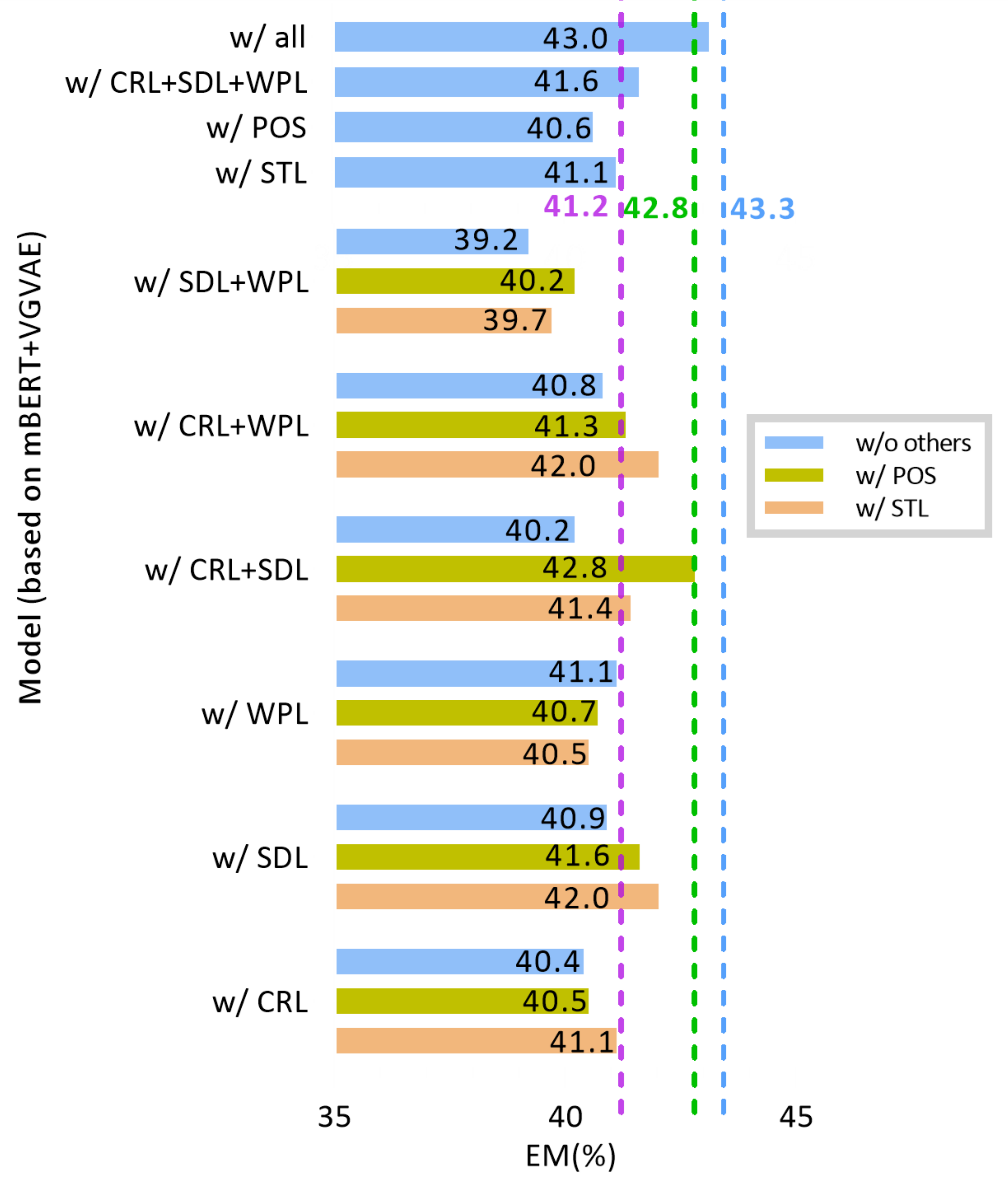}
\caption{The ablation study results on the MLQA dataset. The three dotted lines indicate the results of baseline mBERT, S$^2$DM\_POS, and S$^2$DM\_SP from left to right, respectively. }
\label{ablation}
\end{figure}

\begin{figure}
\centering
\includegraphics[width=6cm]{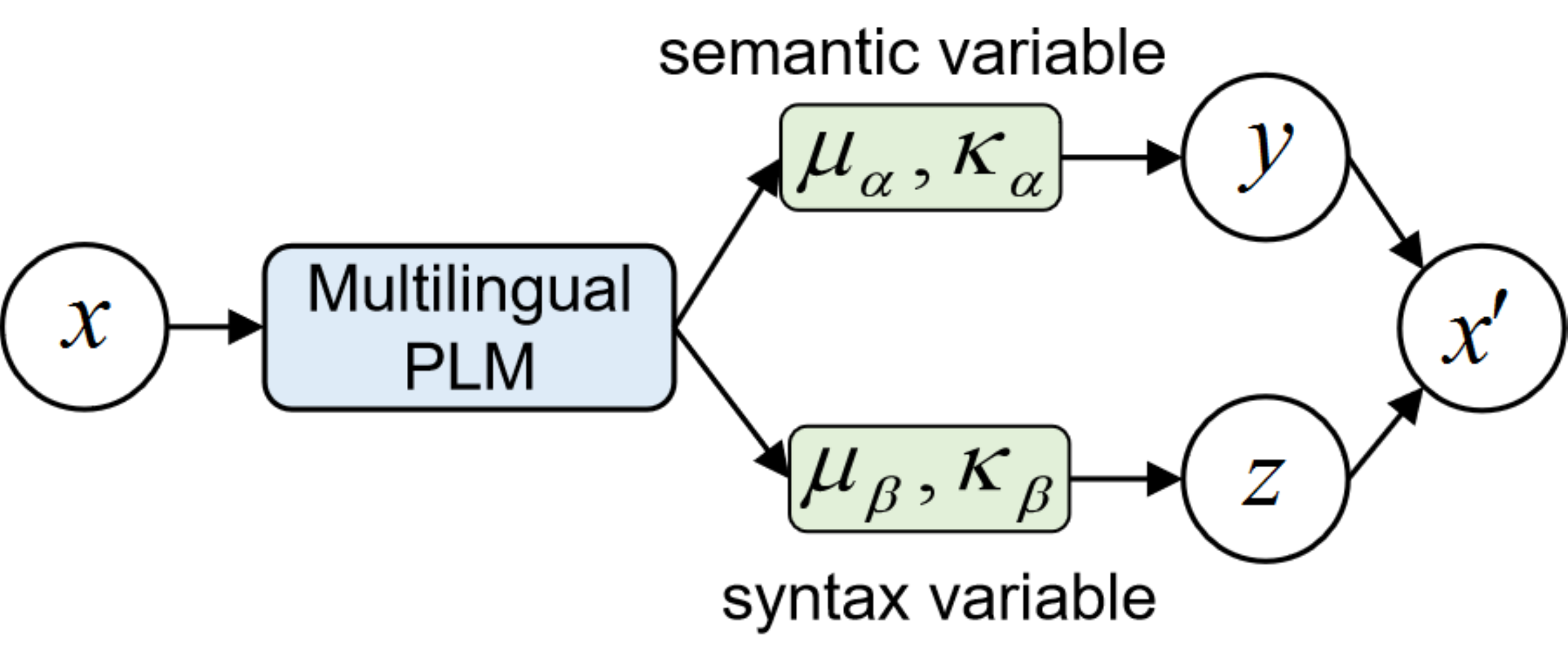}
\caption{A single network of S$^2$DM.}
\label{single}
\end{figure}

We further conducted an ablation study based on the mBERT and VGVAE model with different combinations of losses (introduced in the Section.\ref{ssdm}). The results are shown in Figure~\ref{ablation}. Our mBERT+S$^2$DM\_SP MRC model achieves the strongest performance among all variants, surpassing the model w/ all losses. According to the results shown in Figure~\ref{ablation}, we can summarize that each loss is essential and suitable to our model.

The results without POS and STL loss (e.g., w/ CRL+SDL+WPL) on the MLQA dataset validate the effectiveness of our losses (POS or STL loss) tailored for capturing syntactic information. The performance of models that only contain two losses in CRL, SDL, and WPL drops significantly compared with the w/ CRL+SDL+WPL model. The results of models that only contain one of the losses in CRL, SDL drop slightly, but the EM of the model with only WPL is better than w/ CRL+WPL and w/ SDL+WPL, which further demonstrates the importance of the syntax-oriented loss. All ablation models do not exceed our best model, illustrating the importance of all proposed losses.

\subsection{Why Use a Siamese Network in S$^2$DM?}

\begin{figure*}
\centering
\includegraphics[width=15cm]{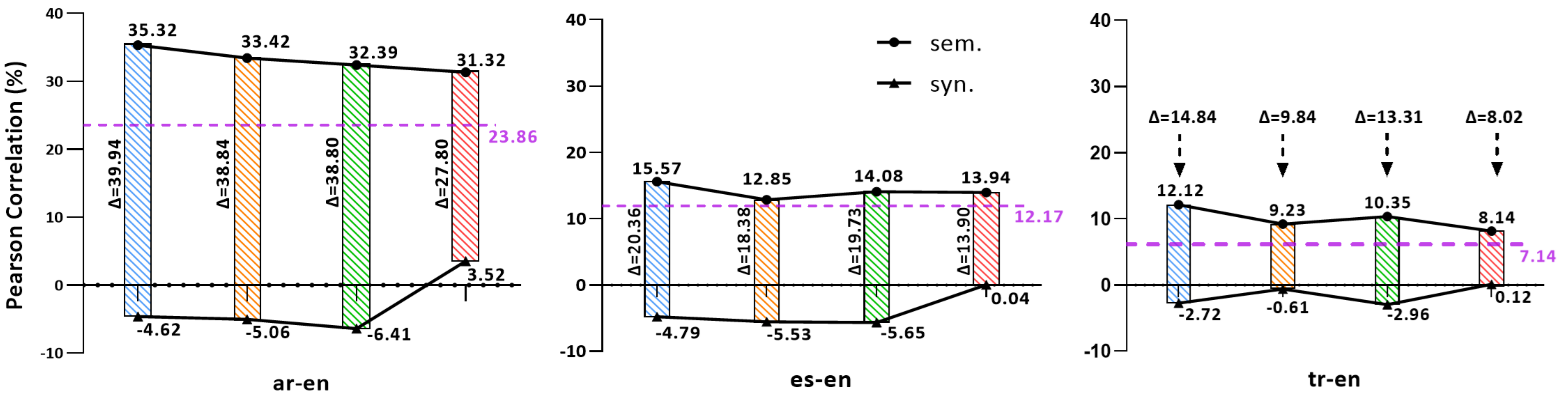}
\caption{Pearson correlation (\%) on the cross-lingual STS tasks. The length of the bar represents the gap of two vectors. The four bars with different colors represent the results of S$^2$DM\_SP, S$^2$DM\_single\_SP, S$^2$DM\_POS, and S$^2$DM\_single\_POS from left to right. Purple dotted line: the result of mBERT. }
\label{sts}
\end{figure*}

\begin{figure}
\centering
\includegraphics[width=7.5cm]{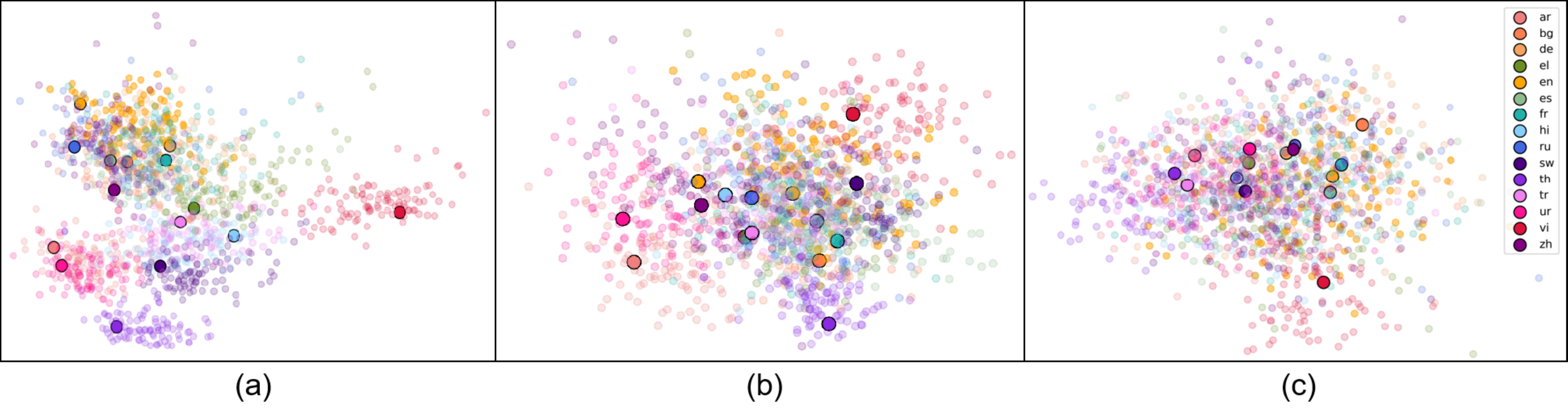}
\caption{PCA visualization of hidden representations from the last layer of mBERT (a) and semantic representations of S$^2$DM\_POS (b) and S$^2$DM\_SP (c). Darker dots: the same 15-way parallel sentence in different languages. }
\label{PCA}
\end{figure}

\begin{table}
\centering
\resizebox{7.5cm}{!}{
\begin{tabular}{l|c|c|c|c|c|c}
\hline
\multirow{2}{*}{} & \multicolumn{2}{c|}{XQuAD}      & \multicolumn{2}{c|}{MLQA}       & \multicolumn{2}{c}{TyDi QA}    \\ \cline{2-7} 
                  & EM             & F1             & EM             & F1             & EM             & F1              \\ \hline
S$^2$DM\_POS         & 49.4          		& 63.7          		& 42.8          		& 59.9          		& 46.3          		& 58.7         \\ 
S$^2$DM\_single\_POS & 48.6          & 62.6          & 42.4          & 59.2          & 43.9          & 56.4          \\ 
S$^2$DM\_SP          & \textbf{49.8} 	& \textbf{64.1} 	& \textbf{43.3} 	& \textbf{60.3} 	&\textbf{47.8}   &\textbf{60.1}   \\
S$^2$DM\_single\_SP  & 49.2         & 63.6          & 42.5          & 59.6          & 45.3          & 58.4      \\ \hline
\end{tabular}}
\caption{Performance of our multilingual MRC model with different S$^2$DM variants based on mBERT.}
\label{single_t}
\end{table}

In order to separate semantic information from PLMs, an alternative way is to train a single network based on the VGVAE model as shown in Figure~\ref{single}. Compared with S$^2$DM, the single-network model does not use the CRL and SDL loss and only requires labeled monolingual data. Corresponding to S$^2$DM, there are also two single-network variants: S$^2$DM\_single\_POS and S$^2$DM\_single\_SP. Since there is no explicit semantics learning across the source and target language, we conjecture that the single-network S$^2$DM will affect the quality of learned semantic vectors and the degree of semantics-syntax decoupling. As shown in Table~\ref{single_t}, the performance of the single-network S$^2$DM is worse than the siamese-network model.

\subsection{Why the S$^2$DM Works?}
Our method mainly aims to reduce the potential negative impact of syntactic differences of languages in the zero-shot transfer process by explicitly isolating semantics from syntax in representations from multilingual pre-trained models. Therefore, we hope to obtain multilingual semantic representations with rich semantic information to guide the machine to read and understand texts. In order to examine (1) whether semantic vectors $y$ in S$^2$DM encode rich semantic information, and (2) whether semantics is sufficiently separated from syntax, and (3) whether semantic disentanglement can improve predicted answer spans in matching syntactic structures of the target language, we conducted additional experiments and analyses.

Here we used three datasets of cross-lingual semantic textual similarity (STS) in SemEval-2017\footnote{\url{https://alt.qcri.org/semeval2017/task1/}} to evaluate the quality of semantic vectors learned by S$^2$DM. The three datasets are for Arabic to English (ar-en), Spanish to English (es-en), and Turkish to English (tr-en) cross-lingual STS. We report the results of our models in Figure~\ref{sts} based on mBERT. We also evaluated learned syntactic vectors in cross-lingual STS, hoping that the performance gap between semantic vectors (i.e., $y$ in S$^2$DM) and syntactic vectors (i.e., $z$ in S$^2$DM) is as large as possible. As shown in Figure~\ref{sts}, disentangled semantic representations significantly improve Pearson correlation over the baseline in ar-en, es-en, and tr-en by 11.46\%, 3.40\%, 4.98\%, respectively. Additionally, disentangled syntactic representations are negatively correlated to STS in most cases. These results suggest that disentangled semantic vectors indeed learn rich universal semantic information.

We visualize hidden representations of the last layer of mBERT and semantic representations of mBERT+S$^2$DM\_POS and mBERT+S$^2$DM\_SP in Figure~\ref{PCA}, in which the parallel sentences are from a 15-way parallel corpus \citep{XNLI}. It is clear to see that disentangled semantic representations learned by S$^2$DM make parallel sentences in 15 languages (semantically equivalent to each other) closer to one another in space, blending language boundaries clearly seen from mBERT representations (Figure~\ref{PCA}(a)). Combined with the negative/positive results of syntactic/semantic vectors in the cross-lingual STS task in SemEval-2017, the visualization demonstrates that S$^2$DM can efficiently disassociate semantics from syntax.

Finally, we evaluated the degree of consistency to syntactic constituents of predicted answer spans. As described in Section~\ref{introduction}, 23.15\% of the non-transfer predicted correct answers violate syntactic constraint of the target language during the raw zero-shot cross-lingual transfer on BiPaR. By contrast, S$^2$DM\_POS and S$^2$DM\_SP drop this percentage to 12.98\% and 6.60\%, respectively. Moreover, on the entire test set of BiPaR ~\citep{BiPaR} in Chinese, 93.27\% answers predicted by S$^2$DM\_SP exactly span syntactic constituents, which is 8.14\% higher than the mBERT model.


\section{Conclusions}
In this paper, we have presented a novel multilingual MRC model for zero-shot cross-lingual transfer, which can disentangle semantic from syntactic representations and explicitly transfer semantic information from rich-resource language to low-resource languages, reducing the influence of syntactic differences between languages on the answer span prediction of the target language. To disassociate semantics from syntax in multilingual pre-trained representations, we propose the siamese semantic disentanglement model that semantics/syntax-oriented losses to guide latent variables to learn corresponding information. For low-resource languages without training data for the decoupling model, our theoretical analysis and experiments verify the generalization of our multilingual MRC model. Further in-depth analyses suggest that the proposed S$^2$DM can efficiently disentangle semantics from syntax and significantly improve syntactic consistency of answer predictions on the target language after zero-shot cross-lingual transfer.

\section*{Acknowledgments}
The present research was supported by the National Natural Science Foundation of China (NSFC) (61972455), the Joint Project of AISHU.com, Bayescom, Zhejiang Lab (No. 2022KH0AB01) and the Natural Science Foundation of Tianjin (No. 19JCZDJC31400). Xiaowang Zhang is supported by the program of Peiyang Young Scholars in Tianjin University (2019XRX-0032). 

\bibliography{custom}

\begin{thebibliography}{25}
\expandafter\ifx\csname natexlab\endcsname\relax\def\natexlab#1{#1}\fi

\bibitem[{Artetxe et~al.(2020)Artetxe, Ruder, and Yogatama}]{XQuAD}
Mikel Artetxe, Sebastian Ruder, and Dani Yogatama. 2020.
\newblock \href {https://doi.org/10.18653/v1/2020.acl-main.421} {On the
  cross-lingual transferability of monolingual representations}.
\newblock In \emph{Proceedings of the 58th Annual Meeting of the Association
  for Computational Linguistics, {ACL} 2020}, pages 4623--4637.

\bibitem[{Chen et~al.(2019)Chen, Tang, Wiseman, and Gimpel}]{ChenTWG19}
Mingda Chen, Qingming Tang, Sam Wiseman, and Kevin Gimpel. 2019.
\newblock \href {https://doi.org/10.18653/v1/n19-1254} {A multi-task approach
  for disentangling syntax and semantics in sentence representations}.
\newblock In \emph{Proceedings of the 2019 Conference of the North American
  Chapter of the Association for Computational Linguistics: Human Language
  Technologies, {NAACL-HLT} 2019}, pages 2453--2464.

\bibitem[{Chi et~al.(2020)Chi, Hewitt, and Manning}]{multiprobe}
Ethan~A. Chi, John Hewitt, and Christopher~D. Manning. 2020.
\newblock \href {https://doi.org/10.18653/v1/2020.acl-main.493} {Finding
  universal grammatical relations in multilingual {BERT}}.
\newblock In \emph{Proceedings of the 58th Annual Meeting of the Association
  for Computational Linguistics, {ACL} 2020}, pages 5564--5577.

\bibitem[{Clark et~al.(2020)Clark, Palomaki, Nikolaev, Choi, Garrette, Collins,
  and Kwiatkowski}]{TyDiQA}
Jonathan~H. Clark, Jennimaria Palomaki, Vitaly Nikolaev, Eunsol Choi, Dan
  Garrette, Michael Collins, and Tom Kwiatkowski. 2020.
\newblock \href {https://transacl.org/ojs/index.php/tacl/article/view/1929}
  {Tydi {QA:} {A} benchmark for information-seeking question answering in
  typologically diverse languages}.
\newblock \emph{Trans. Assoc. Comput. Linguistics}, 8:454--470.

\bibitem[{Conneau et~al.(2020)Conneau, Khandelwal, Goyal, Chaudhary, Wenzek,
  Guzm{\'{a}}n, Grave, Ott, Zettlemoyer, and Stoyanov}]{XLM-R}
Alexis Conneau, Kartikay Khandelwal, Naman Goyal, Vishrav Chaudhary, Guillaume
  Wenzek, Francisco Guzm{\'{a}}n, Edouard Grave, Myle Ott, Luke Zettlemoyer,
  and Veselin Stoyanov. 2020.
\newblock \href {https://doi.org/10.18653/v1/2020.acl-main.747} {Unsupervised
  cross-lingual representation learning at scale}.
\newblock In \emph{Proceedings of the 58th Annual Meeting of the Association
  for Computational Linguistics, {ACL} 2020}, pages 8440--8451.

\bibitem[{Conneau and Lample(2019)}]{XLM}
Alexis Conneau and Guillaume Lample. 2019.
\newblock \href
  {https://proceedings.neurips.cc/paper/2019/hash/c04c19c2c2474dbf5f7ac4372c5b9af1-Abstract.html}
  {Cross-lingual language model pretraining}.
\newblock In \emph{Proceedings of the 2019 Annual Conference on Neural
  Information Processing Systems, NeurIPS 2019}, pages 7057--7067.

\bibitem[{Conneau et~al.(2018)Conneau, Rinott, Lample, Williams, Bowman,
  Schwenk, and Stoyanov}]{XNLI}
Alexis Conneau, Ruty Rinott, Guillaume Lample, Adina Williams, Samuel~R.
  Bowman, Holger Schwenk, and Veselin Stoyanov. 2018.
\newblock \href {https://doi.org/10.18653/v1/d18-1269} {{XNLI:} {E}valuating
  cross-lingual sentence representations}.
\newblock In \emph{Proceedings of the 2018 Conference on Empirical Methods in
  Natural Language Processing, {EMNLP} 2018}, pages 2475--2485.

\bibitem[{Cui et~al.(2019)Cui, Che, Liu, Qin, Wang, and Hu}]{Dual-BERT}
Yiming Cui, Wanxiang Che, Ting Liu, Bing Qin, Shijin Wang, and Guoping Hu.
  2019.
\newblock \href {https://doi.org/10.18653/v1/D19-1169} {Cross-lingual machine
  reading comprehension}.
\newblock In \emph{Proceedings of the 2019 Conference on Empirical Methods in
  Natural Language Processing and the 9th International Joint Conference on
  Natural Language Processing, {EMNLP-IJCNLP} 2019}, pages 1586--1595.

\bibitem[{Devlin et~al.(2019)Devlin, Chang, Lee, and Toutanova}]{BERT}
Jacob Devlin, Ming{-}Wei Chang, Kenton Lee, and Kristina Toutanova. 2019.
\newblock \href {https://doi.org/10.18653/v1/n19-1423} {{BERT:} {P}re-training
  of deep bidirectional transformers for language understanding}.
\newblock In \emph{Proceedings of the 2019 Conference of the North American
  Chapter of the Association for Computational Linguistics: Human Language
  Technologies, {NAACL-HLT} 2019}, pages 4171--4186.

\bibitem[{Hewitt and Manning(2019)}]{structural_probe}
John Hewitt and Christopher~D. Manning. 2019.
\newblock \href {https://doi.org/10.18653/v1/n19-1419} {A structural probe for
  finding syntax in word representations}.
\newblock In \emph{Proceedings of the 2019 Conference of the North American
  Chapter of the Association for Computational Linguistics: Human Language
  Technologies, {NAACL-HLT} 2019}, pages 4129--4138.

\bibitem[{Hsu et~al.(2019)Hsu, Liu, and Lee}]{HsuLL19}
Tsung{-}Yuan Hsu, Chi{-}Liang Liu, and Hung{-}yi Lee. 2019.
\newblock \href {https://doi.org/10.18653/v1/D19-1607} {Zero-shot reading
  comprehension by cross-lingual transfer learning with multi-lingual language
  representation model}.
\newblock In \emph{Proceedings of the 2019 Conference on Empirical Methods in
  Natural Language Processing and the 9th International Joint Conference on
  Natural Language Processing, {EMNLP-IJCNLP} 2019}, pages 5932--5939.

\bibitem[{Hu et~al.(2020)Hu, Ruder, Siddhant, Neubig, Firat, and
  Johnson}]{XTREME}
Junjie Hu, Sebastian Ruder, Aditya Siddhant, Graham Neubig, Orhan Firat, and
  Melvin Johnson. 2020.
\newblock \href {http://proceedings.mlr.press/v119/hu20b.html} {{XTREME:} {A}
  massively multilingual multi-task benchmark for evaluating cross-lingual
  generalisation}.
\newblock In \emph{Proceedings of the 37th International Conference on Machine
  Learning, {ICML} 2020}, pages 4411--4421.

\bibitem[{Hu et~al.(2017)Hu, Yang, Liang, Salakhutdinov, and Xing}]{HuYLSX17}
Zhiting Hu, Zichao Yang, Xiaodan Liang, Ruslan Salakhutdinov, and Eric~P. Xing.
  2017.
\newblock \href {http://proceedings.mlr.press/v70/hu17e.html} {Toward
  controlled generation of text}.
\newblock In \emph{Proceedings of the 34th International Conference on Machine
  Learning, {ICML} 2017}, pages 1587--1596.

\bibitem[{Huang et~al.(2021)Huang, Huang, and Lee}]{self-training}
Wei{-}Cheng Huang, Chien{-}yu Huang, and Hung{-}yi Lee. 2021.
\newblock \href {https://arxiv.org/abs/2105.03627} {Improving cross-lingual
  reading comprehension with self-training}.
\newblock \emph{arXiv preprint arXiv:2105.03627}.

\bibitem[{Jing et~al.(2019)Jing, Xiong, and Zhen}]{BiPaR}
Yimin Jing, Deyi Xiong, and Yan Zhen. 2019.
\newblock \href {https://doi.org/10.18653/v1/D19-1249} {{BiPaR}: {A} bilingual
  parallel dataset for multilingual and cross-lingual reading comprehension on
  novels}.
\newblock In \emph{Proceedings of the 2019 Conference on Empirical Methods in
  Natural Language Processing and the 9th International Joint Conference on
  Natural Language Processing, {EMNLP-IJCNLP} 2019}, pages 2452--2462.

\bibitem[{Lewis et~al.(2020)Lewis, Oguz, Rinott, Riedel, and Schwenk}]{MLQA}
Patrick S.~H. Lewis, Barlas Oguz, Ruty Rinott, Sebastian Riedel, and Holger
  Schwenk. 2020.
\newblock \href {https://doi.org/10.18653/v1/2020.acl-main.653} {{MLQA:}
  {E}valuating cross-lingual extractive question answering}.
\newblock In \emph{Proceedings of the 58th Annual Meeting of the Association
  for Computational Linguistics, {ACL} 2020}, pages 7315--7330.

\bibitem[{Liang et~al.(2021)Liang, Shou, Pei, Gong, Zuo, and
  Jiang}]{CalibreNet}
Shining Liang, Linjun Shou, Jian Pei, Ming Gong, Wanli Zuo, and Daxin Jiang.
  2021.
\newblock \href {https://doi.org/10.1145/3437963.3441728} {Calibrenet:
  Calibration networks for multilingual sequence labeling}.
\newblock In \emph{{WSDM} '21, The Fourteenth {ACM} International Conference on
  Web Search and Data Mining, Virtual Event, Israel, March 8-12, 2021}, pages
  842--850.

\bibitem[{Liu et~al.(2020)Liu, Shou, Pei, Gong, Yang, and Jiang}]{LiuSPGYJ20}
Junhao Liu, Linjun Shou, Jian Pei, Ming Gong, Min Yang, and Daxin Jiang. 2020.
\newblock \href {https://doi.org/10.18653/v1/2020.coling-main.244}
  {Cross-lingual machine reading comprehension with language branch knowledge
  distillation}.
\newblock In \emph{Proceedings of the 28th International Conference on
  Computational Linguistics, {COLING} 2020}, pages 2710--2721.

\bibitem[{Rajpurkar et~al.(2016)Rajpurkar, Zhang, Lopyrev, and
  Liang}]{SQuAD1.0}
Pranav Rajpurkar, Jian Zhang, Konstantin Lopyrev, and Percy Liang. 2016.
\newblock \href {https://doi.org/10.18653/v1/d16-1264} {{SQuAD}: 100, 000+
  questions for machine comprehension of text}.
\newblock In \emph{Proceedings of the 2016 Conference on Empirical Methods in
  Natural Language Processing, {EMNLP} 2016}, pages 2383--2392.

\bibitem[{Tang et~al.(2016)Tang, Bai, and Ma}]{STS}
Shancheng Tang, Yunyue Bai, and Fuyu Ma. 2016.
\newblock \href {https://github.com/IAdmireu/ChineseSTS} {Chinese semantic text
  similarity trainning dataset}.
\newblock Xi'an University of Science and Technology.

\bibitem[{Wu et~al.(2021)Wu, Xu, Qin, Kong, Liu, Zhao, and Chang}]{210705002}
Gaochen Wu, Bin Xu, Yuxin Qin, Fei Kong, Bangchang Liu, Hongwen Zhao, and Dejie
  Chang. 2021.
\newblock \href {https://arxiv.org/abs/2107.05002} {Improving low-resource
  reading comprehension via cross-lingual transposition rethinking}.
\newblock \emph{arXiv preprint arXiv:2107.05002}.

\bibitem[{Yin et~al.(2018)Yin, Zhou, He, and Neubig}]{NeubigZYH18}
Pengcheng Yin, Chunting Zhou, Junxian He, and Graham Neubig. 2018.
\newblock \href {https://www.aclweb.org/anthology/P18-1070/} {{StructVAE}:
  {T}ree-structured latent variable models for semi-supervised semantic
  parsing}.
\newblock In \emph{Proceedings of the 56th Annual Meeting of the Association
  for Computational Linguistics, {ACL} 2018}, pages 754--765.

\bibitem[{Yuan et~al.(2020)Yuan, Shou, Bai, Gong, Liang, Duan, Fu, and
  Jiang}]{YuanSBGLDFJ20}
Fei Yuan, Linjun Shou, Xuanyu Bai, Ming Gong, Yaobo Liang, Nan Duan, Yan Fu,
  and Daxin Jiang. 2020.
\newblock \href {https://doi.org/10.18653/v1/2020.acl-main.87} {Enhancing
  answer boundary detection for multilingual machine reading comprehension}.
\newblock In \emph{Proceedings of the 58th Annual Meeting of the Association
  for Computational Linguistics, {ACL} 2020}, pages 925--934.

\bibitem[{Zeman et~al.(2020)Zeman, Nivre, Abrams et~al.}]{UD2.7}
Daniel Zeman, Joakim Nivre, Mitchell Abrams, et~al. 2020.
\newblock \href {http://hdl.handle.net/11234/1-3424} {Universal dependencies
  2.7}.
\newblock {LINDAT}/{CLARIAH}-{CZ} digital library at the Institute of Formal
  and Applied Linguistics ({{\'U}FAL}), Faculty of Mathematics and Physics,
  Charles University.

\bibitem[{Zhang et~al.(2019)Zhang, Yang, Yuan, Shen, and Carin}]{ZhangYYSC19}
Xinyuan Zhang, Yi~Yang, Siyang Yuan, Dinghan Shen, and Lawrence Carin. 2019.
\newblock \href {https://doi.org/10.18653/v1/p19-1199} {Syntax-infused
  variational autoencoder for text generation}.
\newblock In \emph{Proceedings of the 57th Conference of the Association for
  Computational Linguistics, {ACL} 2019}, pages 2069--2078.

\end{thebibliography}
\bibliographystyle{acl_natbib}



\end{document}